\documentclass[11pt,a4paper]{article}
\usepackage[hyperref]{emnlp2020}
\usepackage{times}
\usepackage{latexsym}
\usepackage{float}
\usepackage[justification=centering]{caption}
\usepackage{url}
\usepackage{algorithm, algorithmic}
\usepackage{graphicx}
\usepackage{multirow}
\usepackage{array}
\usepackage{dblfloatfix}
\usepackage{url}
\usepackage{amsmath}
\usepackage{enumitem}
\usepackage{subcaption} 
\newsavebox{\measurebox}

\usepackage{tabularx}
\usepackage{caption}

\usepackage{microtype}

\aclfinalcopy 
\def\aclpaperid{1129} 

\title{Chapter Captor: Text Segmentation in Novels}

\author{Charuta Pethe, Allen Kim, Steven Skiena \\ Department of Computer Science, \\ Stony Brook University, NY, USA \\ \texttt{\{cpethe,allekim,skiena\}@cs.stonybrook.edu}}

\date{}

\begin{document}
\maketitle
\begin{abstract}
Books are typically segmented into chapters and sections, representing coherent sub-narratives and topics. We investigate the task of predicting chapter boundaries, as a proxy for the general task of segmenting long texts.
We build a Project Gutenberg chapter segmentation data set of 9,126 English novels, using a hybrid approach combining neural inference and rule matching to recognize chapter title headers in books, achieving an F1-score of 0.77 on this task. Using this annotated data as ground truth after removing structural cues, we present cut-based and neural methods for chapter segmentation, achieving an F1-score of 0.453 on the challenging task of \textit{exact} break prediction over book-length documents.
Finally, we reveal interesting historical trends in the chapter structure of novels.
\end{abstract}

\section{Introduction}

Text segmentation \cite{hearst_1994, beeferman_1999} is a fundamental task in natural language processing, which seeks to partition texts into sequences of coherent segments or episodes.  Segmentation tasks differ widely in scale, from partitioning sentences into clauses to dividing large texts into coherent parts, where each segment is ideally an independent event occurring in the narrative.

Text segmentation plays an important role in many NLP applications including summarization, information retrieval, and question answering.
In the context of literary works, event detection is a central concern in discourse analysis \cite{discourse_2019}. In order to obtain representations of events, it is essential to identify narrative boundaries in the text, where one event ends and another begins.

In novels and related literary works, authors often define such coherent segments by means of sections and chapters. Chapter boundaries are typically denoted by formatting conventions such as page breaks, white-space, chapter numbers, and titles.  This physical segmentation improves the readability of long texts for human readers, providing transition cues for breaks in the story.

In this paper, we investigate the task of identifying chapter boundaries in literary works, as a proxy for that of large-scale text segmentation.   The text of thousands of scanned books are available in repositories such as Project Gutenberg \cite{project_gutenberg}, making the chapter boundaries of these texts an attractive source of annotations to study text segmentation.   Unfortunately, the physical manifestations of the printed book have been lost in the Gutenberg texts, limiting their usefulness for such studies.
Chapter titles and numbers are retained in the texts but not systematically annotated: indeed they sit as hidden obstacles for most NLP analysis of these texts.

We develop methods for extracting ground truth chapter segmentation from Gutenberg texts, and use this as training/evaluation data to build text segmentation systems to predict the natural boundaries of long narratives. Our primary contributions \footnote{All code and links to data are available at \url{https://github.com/cpethe/chapter-captor}.} include:

\begin{itemize}
    \item \textbf{Project Gutenberg Chapter Segmentation Resource:} To create a ground-truth data set for chapter segmentation, we developed a hybrid approach to recognizing chapter formatting which is of independent interest.  It combines a neural model with a regular expression based rule matching system. Evaluation on a (noisy) silver-standard chapter partitioning yields a mean value F1 score of 0.77 of a test set of 640 books, but manual investigation shows this evaluation receives an artificially low recall score due to incorrect header tags in the silver-standard.
    
    Our data set consists of 9,126 English fiction books in the Project Gutenberg corpus. To encourage further work on text segmentation for narratives, we make the annotated chapter boundaries data publicly available for future research.

    \item \textbf{Local Methods for Chapter Segmentation:} By concatenating chapter text following the removal of all explicit signals of chapter boundaries (white space and header notations), we create a natural test bed to develop and evaluate algorithms for large-document text segmentation.
    We develop two distinct approaches for predicting the location of chapter breaks: an unsupervised weighted-cut approach minimizing cross-boundary cross-references, and a supervised neural network building on the BERT language model \cite{devlin_2019}.  Both prove effective at identifying likely boundary sites, with F1 scores of 0.164 and 0.447 respectively on the test set.
    
    \item \textbf{Global Break Prediction using Optimization:}  Social conventions encourage authors to maintain chapters of modest yet roughly equal length.  
    By incorporating length criteria into the desired optimization criteria and using dynamic programming to find the best global solution enables us to control how important it is to keep the segments equal. We find that a balance between equal segments and model-influenced segments gives us the best segmentation, with minimal error. Indeed, augmenting the BERT-based local classifier with dynamic programming yielded an F1 score of 0.453 on the challenging task of {\em exact} break prediction over book-length documents, while simultaneously beating challenging baselines on two other error metrics.
    
    Incorporating chapter length criteria require an independent estimate of the number of chapters in a given text. We demonstrate that there are approximately five times as many likely break candidates as there are chapter breaks in the weighted cut approach, reflecting the number of sub-events within an average book chapter.

    \item \textbf{Historical Analysis of Segmentation Conventions} -- We exploit our data analysis of segmented books in two directions.   We demonstrate that novels grew in length to an average of roughly 30 chapters/book by 1800, and retained this length until 1875 before beginning a steady decline.   Second, an analysis of regular expression patterns reveal the wide variety of chapter header conventions and which forms dominate.
    
\end{itemize}

\section{Previous Work}

Many approaches have been developed in recent years to address variants of the task of identifying structural elements in books.

\newcite{mcconnaughey_2017} attempt this task at the page-level, by assigning a label (e.g. Preface, Index, Table of Contents, etc.) to each page of the book. \newcite{wu_2013} address the task of recognizing and extracting tables of contents from book documents, with a focus on identifying its style. Participants of the Book Structure Extraction competition at ICDAR 2013 \cite{doucet_2013} attempted to use various approaches for the task. These include making use of the table of contents, OCR information, whitespace, and indentation.
\newcite{dejean_2005} present approaches to identify a table of contents in a book, and \newcite{dejean_2009} attempt to structure a document according to its table of contents.

However, our approach relies only on text, and does not require positional information or OCR coordinates to extract front matter and headings from book texts.

For text segmentation, many approaches have been developed over the past years, suitable for different types of data, such as news articles, scientific article, Wikipedia pages, and conversation transcripts.

The TextTiling algorithm \cite{hearst_1994} makes use of lexical frequency distributions across blocks of a fixed number of words. Dotplotting \cite{reynar_1994} is a graphical technique to locate discourse boundaries using lexical cohesion across the entire document.

\newcite{yamron_1998} and \newcite{beeferman_1999} propose methods to identify story boundaries in news transcripts.

The C99 algorithm \cite{choi_2000} uses a global lexical similarity matrix and a ranking scheme for divisive clustering. \newcite{choi_2001} further proposed the use of Latent Semantic Analysis (LSA) to compute inter-sentence similarity.

\newcite{utiyama_2001} proposed a statistical model to find the maximum probability segmentation. The Minimum Cut model \cite{barzilay_2006} addresses segmentation as a graph partitioning task.

This problem has also been addressed in a Bayesian setting \cite{eisenstein_2008, eisenstein_2009}. TopicTiling \cite{riedl_2012} is a modification of the TextTiling algorithm, and makes use of LDA for topic modeling.

Segmentation using sentence similarity has been extensively explored using affinity propagation \cite{kazantseva_2011, sakahara_2014}. More recent approaches \cite{alemi_2015, glavavs_2016} involve the use of semantic representations of words to compute sentence similarities.
\newcite{koshorek_2018} and \newcite{badjatiya_2018} propose neural models to identify break points within the text.

\newcite{sims_2019} address the slightly different, but relevant task of event prediction using a neural model, on a human-annotated dataset of short events.

\section{Header Annotation}

In order to create a ground-truth dataset for chapter segmentation, we first build a system to recognize chapter headings, using a hybrid approach combining a neural model with a regular expression (regex)-based rule matching system.

\subsection{Data}
In the absence of human-annotated gold standard data with annotated front matter and chapter headings, we derive silver-standard ground truth from Project Gutenberg. We identify 8,400 English fiction books available in HTML format, and extract (noisy) HTML header elements from these books. We use a train-test split of 90-10\%.

\subsection{Methodology}
\begin{figure}[htbp]
\includegraphics[width=\linewidth]{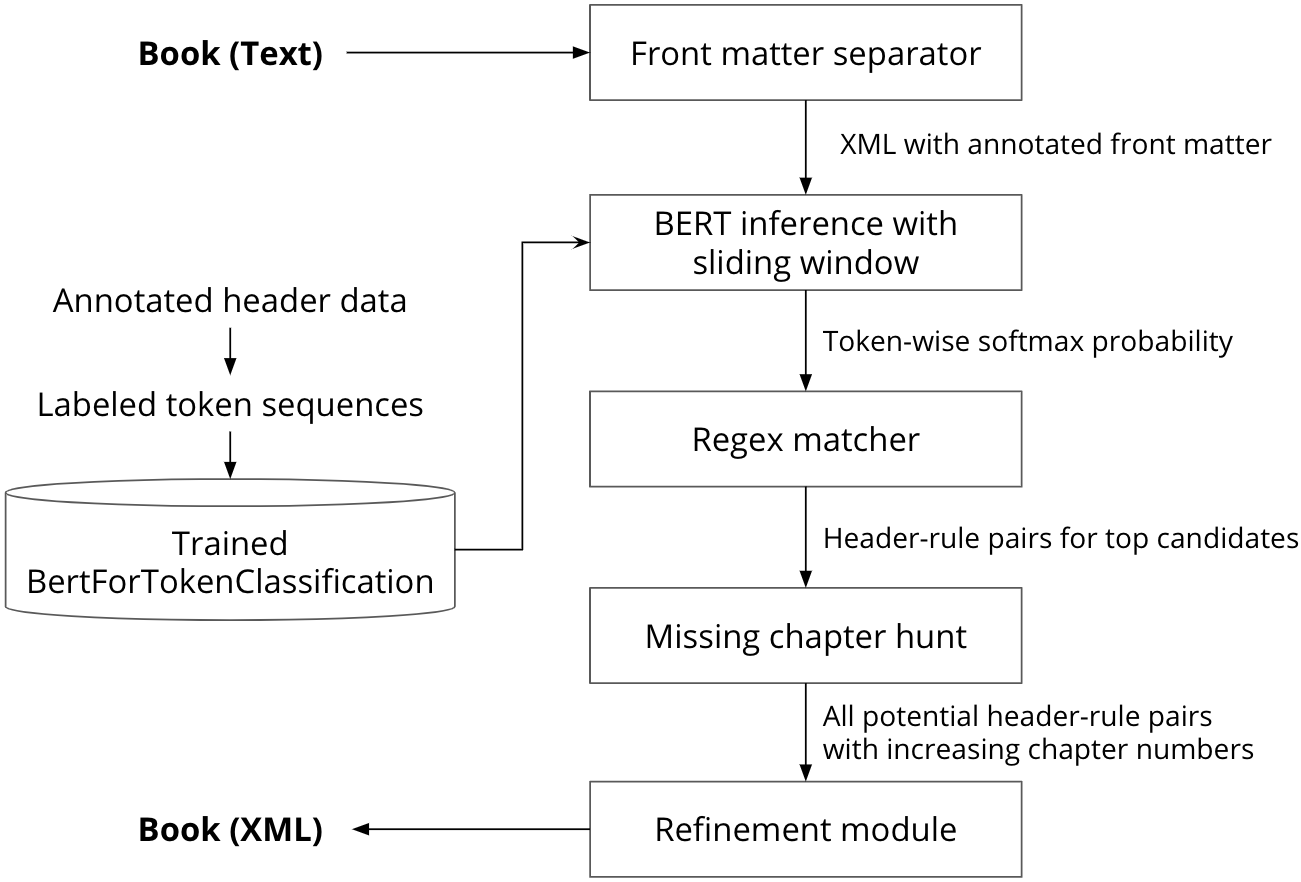}
\caption{Header Annotation Pipeline}
  \label{fig:header_annot_pipeline}
\end{figure}

The annotation pipeline has five components, as shown in Figure \ref{fig:header_annot_pipeline}.
First, we
make use of white-space cues and string matching for keywords such as `Preface', `Table of contents' etc. to identify front matter. We tag all such content up to the first chapter heading as the front matter, and identify the remaining content as body.

\subsubsection{BERT Inference}

We fine-tune a pretrained BERT model \cite{devlin_2019} with a token classification head, to identify the lines which are likely to be headers.

\paragraph{Training:} For each header extracted from the Project Gutenberg HTML files, we append content from before and after the header, to generate training sequences of fixed length. We empirically select a sequence length of $120$. We use a custom BERT Cased Tokenizer with a special token for the newline character, to tokenize the input sequences.
The training samples are of the format:

Sequence: $[p_1,...,p_x, h_1, ..., h_k, q_1, ..., q_y]$

Labels:\hspace{5.8mm}$[0, ......., 0, 1, ......., 1, 0, ......., 0]$

\noindent where $p_1, ..., p_x$ are $x$ tokens before the header, $h_1, ..., h_k$ are $k$ tokens from the header, and $q_1, ..., q_y$ are $y$ tokens after the header. $x$ and $y$ are randomly generated numbers, such that $x + k + y = 120$. This is done in order to prevent header tokens from appearing only in the center of the input sequence.

We fine-tune a pre-trained model for token classification using headers from 6,515 books in our training set for 4 epochs using the BertAdam optimizer. A compute server with a 2.30 GHz CPU and TeslaV100 GPU was used for all experiments.

\paragraph{Inference:} For inference on a test set example, we tokenize the text using the custom BERT Cased Tokenizer, and use the model to generate a confidence score for each token. We do this using a sliding window approach, wherein we run inference on a text window of 120 tokens, and slide the window forward by 60 tokens in each iteration. We then perform token-wise max pooling to obtain a single confidence score per token. Further, we detokenize the output by concatenating sub-word tokens and mean-pooling their confidence scores.

We choose the top 10\% tokens with the highest confidence scores, and use the lines containing these tokens as potential header candidates for regex matching.

\subsubsection{Regex Rule Matching}
We compile a list of regular expressions for constituent elements in chapter headings:
\begin{itemize}[topsep=5pt]
\itemsep0em 
    \item Keywords like `Chapter', `Section', `Volume'
    \item Punctuation marks and whitespace
    \item Title (uppercase and mixed case)
    \item Roman numerals (uppercase and lowercase)
    \item Cardinal, ordinal, and digital numbers.
\end{itemize}

Using the rules for these constituent elements, we further generate a list of 1,015 regex rules for valid permutations of these elements.

For every potential header candidate generated using the BERT model, we pick the best matching regex rule as the longest rule that captures constituent elements in order of priority, and discard the candidate if there is no matching rule.

\subsubsection{Missing Chapter Hunt}
Once we have the list of candidates and their corresponding matching rules, we search for chapter headings the BERT model may have missed. For each matched rule that contains a number in some format, we search for chapter headings in the same format with the missing number. In order to account for chapter numbering restarts in different sections of the book, we search for missing headers within all increasing subsequences in the list of chapter numbers found.

\subsubsection{Refinement}
We get rid of false positive matches, by removing headers between consecutive chapter numbers, which do not match the same rule.

\subsection{Evaluation}

Table \ref{tab:stagewise} shows the stage-wise performance of the annotation pipeline. Stage 1 contains all candidates generated using the BERT model, Stage 2 contains headers predicted after applying regex rules and searching for missing chapters, Stage 3 contains headers after removing false positives.
\begin{table}[htbp]
\centering
\vspace{2mm}
\begin{tabular}{|r|r|r|r|}
\hline
\textbf{Stage} & \textbf{Precision} & \textbf{Recall} & \textbf{F1} \\ \hline
1              & 0.02               & 0.67            & 0.05        \\ \hline
2              & 0.75               & 0.79            & 0.76        \\ \hline
3              & 0.78               & 0.78            & 0.77        \\ \hline
\end{tabular}
\caption{Stage-wise performance for header annotation }
\label{tab:stagewise}
\end{table}

\begin{figure}[htbp]
\centering
\includegraphics[width=0.9\linewidth]{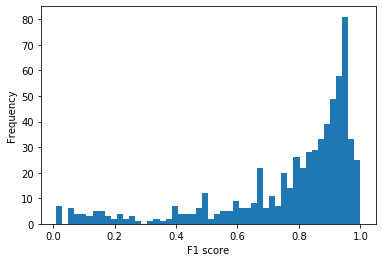}
\caption{F1 score distribution for 640 test set books}
  \label{fig:header_annot_perf}
\end{figure}

Figure \ref{fig:header_annot_perf} shows the distribution of evaluation metrics on the test set of 640 books, evaluated on the ground truth extracted from HTML files. The mean value of the F1 score is $0.77$. Manual investigation of a sub-sample of the test set shows that several books get a low recall score due to false negatives, caused due to incorrect header tags in the silver-standard ground truth.
Thus we have even greater confidence in our testbed than the F1-score suggests.

\subsection{Popularly used rule formats}
\begin{figure}[h]
\includegraphics[width=\linewidth]{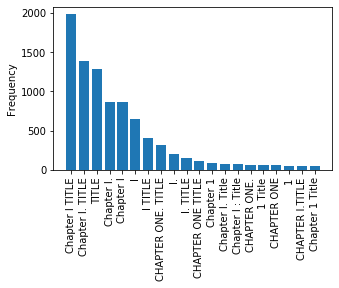}
\caption{Number of books in which the most frequent header formats occur the most frequently}
  \label{fig:rules_hist}
\end{figure}

For each book, we count the number of occurrences of each header format. Figure \ref{fig:rules_hist} shows the number of books in which the respective header format occurs most frequently, namely ``Chapter \# TITLE''.

\subsection{Historical Trends}
\begin{figure}[h]
\includegraphics[width=\linewidth]{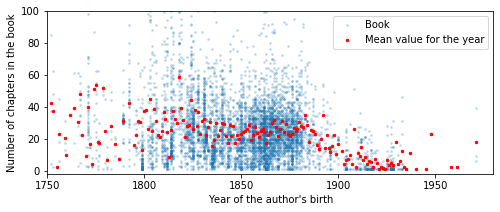}
\caption{Trend in the number of chapters in a book}
  \label{fig:num_chapters_year}
\end{figure}

Figure \ref{fig:num_chapters_year} presents the number of chapters in each book as obtained by our annotation pipeline, against the author's year of birth. For authors born before 1875, novels were roughly 30 chapters long, after which there has been a steady decline in the number of chapters per book.

\begin{figure*}[t]
\centering
  \begin{subfigure}[b]{0.9\textwidth}
  \centering
    \includegraphics[width=\textwidth]{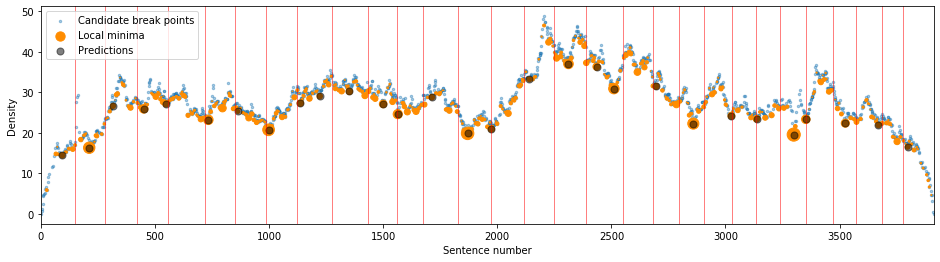}
    \caption{WOC density (Local minima point sizes are proportional to their prominences)}
    \label{fig:woc_graph}
  \end{subfigure}
  \begin{subfigure}[b]{0.9\textwidth}
  \centering
    \includegraphics[width=\textwidth]{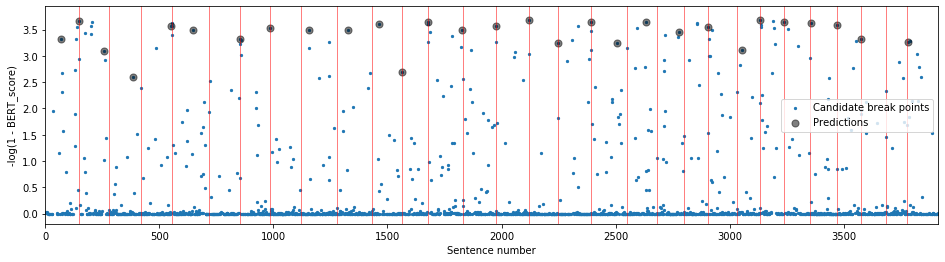}
    \caption{Processed BERT confidence scores}
    \label{fig:bert_graph}
  \end{subfigure}
  \caption{Breakpoint probability scores as a function of sentence number, for a sample book ({\em ``The Rover Boys Out West''}, by Edward Stratemeyer). Vertical red lines denote chapter breaks in ground truth. Predictions are computed using the dynamic programming approach (described in Section \ref{sec:global}) with $\alpha = 0.8$. }
\end{figure*}

\section{Local Methods for Segmentation}

After removing all explicit signals of chapter boundaries from the texts, we now evaluate algorithms for segmenting text into chapters.

We formulate our task as follows:

\noindent\textbf{Given:} Sentences $S_0, S_1, ..., S_{N-1}$ in the book, and $P$, the number of breaks to insert

\noindent\textbf{Compute:} $P$ break points \newline $B_0, B_1, ..., B_{P-1} \in \{0, 1, ..., N-1\}$ corresponding to chapter breaks.

\subsection{Weighted Overlap Cut (WOC)}

The motivation behind this technique is based on the intuition that chapters are relatively self-contained in the words that they use. For example, consider a chapter that refers to a “cabin in the woods”. We would expect references to this cabin to be higher within the same chapter as compared to other chapters. Hence, our hypothesis is that there will be fewer words in common across a break point separating two chapters, as compared to words within the same chapter.

Considering sentences as nodes, and common words as edges, we can compute the density of a potential break point as the sum of the number of edges going across it, weighted by their distance from the break point. As per our hypothesis, we expect the break point between two chapters to appear as local minima in density as a function of sentence number.

We restrict potential break points to the points between paragraphs, and compute the local minima in density. For each local minimum, we compute its prominence as the vertical distance between the minimum and its highest contour line. We then pick the top $P$ most prominent local minima as the break points.

Note that the same hypothesis can also be made at the paragraph level. However, a major limitation of this approach is that paragraph sizes vary widely, ranging from a single word to a considerably huge block of text. Hence, we have taken the approach of computing sentence-level density and then restricting the potential break points to points between paragraphs.

\paragraph{Preprocessing:} We use the Stanford CoreNLP pipeline \cite{stanford_corenlp} for sentence tokenization and lemmatization. We consider paragraphs as text separated by two or more newline characters.

\paragraph{Computation:} For every potential break point $i$ between sentences $S_i$ and $S_{i+1}$, we compute the density of the break point, which is essentially a weighted sum of the number of overlapping word lemmas within a certain window before and after the break point (weighted by the distance of the word occurrence from the break point). We compute the density $d_i$ of break point candidate $i$ as:

\begin{center}
$d_i = \overset{i}{\underset{x=i-w}{\sum}} \left ( \overset{x+w}{\underset{y=\textrm{max}(x+1,i)}{\sum}}\frac{\textrm{overlap}_{xy}}{\left | i - x \right | \left | i - y \right |}   \right )$
\end{center}

where $w$ is the window size and $\textrm{overlap}_{xy}$ is the number of common lemmas in sentences $S_x$ and $S_y$, excluding stopwords and punctuation. (Note that we use only valid sentence indices during summation, considering the first and last sentences of the book as cutoffs.)

\paragraph{Experiments:} We perform experiments on 2,546 books in the test set, using window sizes of 50, 100, 150, and 200 sentences.

Figure \ref{fig:woc_graph} shows the computed densities and local minima for window size 200, for a sample book ({\em ``The Rover Boys Out West''}, by Edward Stratemeyer). The figure shows that chapter breaks roughly correspond to prominent local minima in density.

\subsection{BERT for Break Prediction (BBP)}

We fine-tune a pre-trained BERT model for the Next Sentence Prediction task, to classify pairs of sequences in which the second sequence is a coherent continuation of the first. Intuitively, for text sequences which are separated by a chapter break, we expect the second sequence to not be a continuation of the first, i.e. the output label should be 0. Whereas for consecutive text sequences within the same chapter, the output label should be 1, denoting that it is a logical continuation.

\paragraph{Training:} 
We generate training sequences from 7,582 books in the training set. We generate training examples in the following format:\newline
\texttt{[CLS]<Seq A>[SEP]<Seq B>[SEP]}

To generate negative training samples (i.e. class 0, meaning chapter break), we consider all the chapter boundaries, and construct the input using the text just before the chapter break as \texttt{Seq A}, and text just after the break as \texttt{Seq B}.
To generate positive training samples (i.e. class 1, meaning no break) we consider the break points between paragraphs within the same chapter, and construct the input sequence similarly. We use these sequence pairs to fine-tune a pre-trained model for next sentence prediction. Note that class 0 is of interest to us in this task, as lack of continuity between the sequences denotes the possibility of a chapter break.

\paragraph{Inference:}
During inference on a book, we consider all break points between paragraphs, and generate input sequences as described above. We run each pair of input sequences through the classifier, and generate confidence scores per class. We then use the confidence score for class 0 as the probability of a break. We select the top $P$ break points with the highest confidence scores.

\paragraph{Experiments:} We perform experiments using the following variants of training sequences to fine-tune the BERT model:
\begin{itemize}[topsep=5pt]
\itemsep0em 
    \item \textit{Single paragraph:} We use only one paragraph from before, and one paragraph from after the break point.
    \item \textit{Full window:} We use 254 tokens each, from before and after the break point. (If the paragraph length exceeds 254 tokens, we cut off the text before/after that point, depending on which side the paragraph lies.)

\end{itemize}

Figure \ref{fig:bert_graph} shows the modified BERT scores for the full window configuration, for a sample book in the test set. The figure shows that BERT is able to capture points close to chapter breaks in most cases, indicating a good recall as well as precision.

\subsection{Evaluation}

We evaluate our algorithms using three metrics:

 \paragraph{$\mathbf{P_k}$} \cite{beeferman_1999}: To compute this metric, $k$ is set to half of the average true segment size. Using a moving window of length $k$, a penalty is computed based on whether the two ends of the window are in the same or different segments, and whether the ground truth segmentation is in agreement.
 \paragraph{WindowDiff (WD)} \cite{pevzner_2002}: This metric also uses a moving window, and compares the number of ground truth segmentation boundaries that fall in the window, with the number of boundaries assigned by the algorithm. A penalty is added if the counts are not equal.
 \paragraph{F1 score}: We use the F1 score to evaluate \textit{exact} break prediction, and consider a match only if the break matches with the ground truth exactly, i.e. predictions near the true break points are not counted.

Lower values of $P_k$ and WindowDiff, and a high value for F1 score are indicative of better performance.

Table \ref{tab:no_dp_eval} shows the evaluation metrics $P_k$, WD (WindowDiff), and the F1 score for the WOC and BBP configurations described above.

\setlength{\tabcolsep}{4.5pt}
\vspace{2mm}
\begin{table}[htbp]
\centering
\begin{tabular}{|r|rr|r|}
\hline
\textbf{Algorithm} & $\mathbf{P_k}$ & \textbf{WD} & \textbf{F1} \\ \hline
Equidistant breaks       & 0.482       & 0.492       & 0.052       \\ \hline
TextTile \cite{hearst_1994}               & 0.587       & 0.714       & 0.085       \\
C99 \cite{choi_2000}               & 0.493       & 0.517       & 0.049       \\
P\cite{badjatiya_2018}        & 0.485       & 0.555       & 0.111       \\
L\cite{badjatiya_2018}              & 0.493       & 0.569       & 0.087       \\ \hline
WOC (window=50)    & 0.442       & 0.465       & 0.144       \\
WOC (window=100)   & 0.425       & 0.450       & 0.158       \\
WOC (window=150)   & 0.418       & 0.447       & 0.162       \\
WOC (window=200)   & 0.416       & 0.446       & 0.164       \\ \hline
BBP (single para.) & 0.454       & 0.509       & 0.126       \\
BBP (full window)  & \textbf{0.303}       & \textbf{0.384}       & \textbf{0.447}       \\ \hline
\end{tabular}
\caption{Evaluation metrics for chapter break insertion approaches (For $P_k$, WD: lower is better. For F1: higher is better.)}
\label{tab:no_dp_eval}
\end{table}
\setlength{\tabcolsep}{5pt}

We compare our approaches against the following baselines:
\begin{itemize}[topsep=5pt]
\itemsep0em 
    \item \textbf{Equidistant:} We divide the book into $P+1$ segments, such that each segment has the same number of sentences.
    \item \textbf{TextTiling:} We run the TextTiling algorithm \cite{hearst_1994}, using mean words per sentence as pseudosentence size, and number of paragraphs as block size for each book. The average number of breaks per book inserted by this algorithm is 574, which clearly does not reflect the actual number of chapters, resulting in poor performance.
    \item \textbf{C99:} We run the C99 algorithm \cite{choi_2000} on our dataset, and choose the first $P$ breaks obtained while performing divisive clustering.
    \item \textbf{Perceptron (P): } We train a 3-layer baseline perceptron model with 300 neurons in each layer, for 10 epochs, as described by \newcite{badjatiya_2018}. We use mean-pooled 300-dimensional word2vec embeddings \cite{mikolov_2013} trained on the Google News dataset, as input to the perceptron.
    \item \textbf{LSTM (L): } We train a neural model as described by \newcite{badjatiya_2018}, using the same pre-trained word2vec embedding matrix. The network consists of an Embedding layer, followed by an LSTM layer, a dropout layer, a dense layer and finally, a sigmoid activation layer.
\end{itemize}

Our models outperform the baselines on all metrics, with the BERT (full window) model for break prediction model giving the best results.
The approaches by \newcite{reynar_1994} and \newcite{utiyama_2001}, and the neural models proposed by \newcite{badjatiya_2018} and \newcite{koshorek_2018} are global models, and are prohibitively expensive on long documents.

\section{Global Break Prediction}
\label{sec:global}

In the approaches described above, we simply select the highest scoring $P$ points. However, this selection does not conform to spatial constraints. For example, the model may place two breaks close to one another, when realistically, chapter breaks are spaced fairly apart in practice. 

\begin{figure}[htbp]
\vspace{2mm}
\begin{center}
\includegraphics[width=0.75\linewidth]{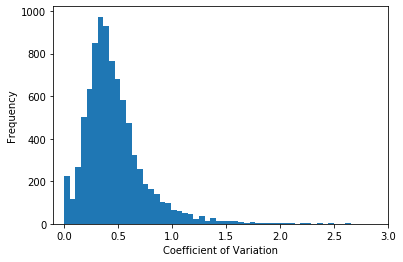}
\end{center}
\caption{Frequency of distribution for the coefficient of variance of number of sentences per segment}
  \label{fig:coeff_var}
\end{figure}

To validate this, we compute the coefficient of variance (CV) for each book, in terms of the number of sentences per chapter. Figure \ref{fig:coeff_var} shows the distribution of the CV over all books in our dataset. Most books in our dataset have a low CV (distributions with CV less than $1$ are considered to be low-variance), reflecting the fact that chapters breaks are spaced fairly equally apart.

Hence, we propose a dynamic programming approach, in order to incorporate a weight for keeping the chapter breaks equidistant.

We formulate the task in the same way as described previously, with an additional parameter $\alpha$, which determines the importance of the confidence scores as compared to equidistant breaks. $\alpha$ ranges from $0$ to $1$, where $1$ indicates that full weight is given to confidence scores, and $0$ indicates that full weight is given to keeping the breaks equidistant. We define the cost of inserting a break at point $n$ and inserting $k$ breaks in points $0$ to $n - 1$ recursively as:
\newline
\newline
\resizebox{\linewidth}{!}{
    $\textrm{cost}(n,k) =\underset{i\in\left [0,n-1  \right ]}{\textrm{min}}\left (  \textrm{cost}(i,k-1) + (1 - \alpha)\frac{ \left | n - i\right |}{L}\right ) - \alpha \cdot s_n$
}
\newline
\newline
where $s_n$ is the confidence score for $n$ being a break point, and $L$ is the ideal chapter length, i.e. number of sentences in each chapter if the book is split into $P+1$ equal parts. At each step, we use the break point which results in cost minimization as the next break point, and repeat the recursive call.

\subsection{Experiments}

We apply dynamic programming for global break prediction, to both the approaches described above. We conduct experiments for $\alpha$ from $0$ to $1$, with a step increase of $0.2$.

\subsubsection{WOC}
We use the prominences of local minima obtained using WOC, with window sizes 50, 100, 150, and 200 respectively. We apply min-max normalization on the prominences.

\begin{figure}[htbp]
\begin{center}
\includegraphics[width=0.875\linewidth]{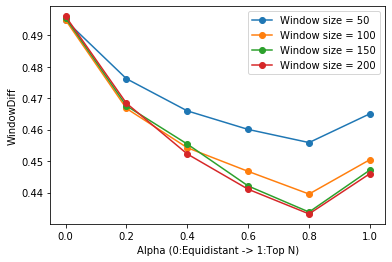}
\end{center}
\caption{WindowDiff error metric for WOC}
  \label{fig:dp_woc}
\end{figure}

Figure \ref{fig:dp_woc} shows the WindowDiff error metric for the WOC approach, with differing window sizes, for different values of $\alpha$. (Note that the window sizes here are in terms of the number of neighboring sentences used to compute density, and not used while calculating the WindowDiff metric.) The figure shows that an increase in window size results in lower error, and for all window sizes, $\alpha=0.8$ shows the best performance.

\subsection{BBP}

We use the confidence scores for class 0 obtained using the BERT model. We observe that confidence scores are clustered close to $0$ and $1$. Higher confidence scores are of more interest to us, as they are indicative of potential chapter boundaries. Hence, in order to distribute the values closer to $1$ further apart, we apply the log function and compute the modified confidence score as $-\ln(1 - \textrm{score}) / c$, where $c$ is a normalizing constant. In practice, we use $c=10$ to limit a majority of the values between $0$ and $1$. We optimize for the best value of alpha independently of this constant.

\begin{figure}[htbp]
\begin{center}
\includegraphics[width=0.875\linewidth]{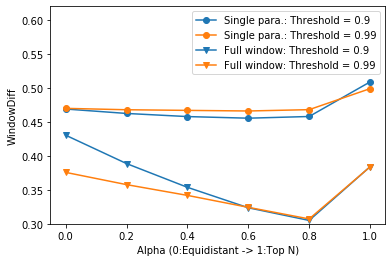}
\end{center}
\caption{WindowDiff error metric for BBP}
  \label{fig:dp_bert_both}
\end{figure}

Figure \ref{fig:dp_bert_both} shows the WindowDiff error metric for the BERT-based approach for the single paragraph and full window models respectively. We use thresholds of $0.9$ and $0.99$ for each of the models, meaning that we consider only those break points with confidence scores above the threshold as potential break point candidates.

The full window model shows the least error at $\alpha=0.8$. Note that a higher threshold of $0.99$ shows a performance almost equal to that of $0.9$, since a higher threshold means fewer potential break point candidates, and hence a lower runtime. Figures \ref{fig:woc_graph} and \ref{fig:bert_graph} depict predictions from the WOC and BBP approaches respectively, with $\alpha=0.8$.

\begin{table}[htbp]
\vspace{2mm}
\centering
\begin{tabular}{|r|rr|r|}
\hline
\textbf{Algorithm} & $\mathbf{P_k}$ & \textbf{WD} & \textbf{F1} \\ \hline
Best BBP (local)    & 0.303       & 0.384       & 0.447       \\ \hline
WOC (window=50)    & 0.443       & 0.456       & 0.144       \\
WOC (window=100)   & 0.426       & 0.440       & 0.158       \\
WOC (window=150)   & 0.421       & 0.434       & 0.162       \\
WOC (window=200)   & 0.420       & 0.433       & 0.164       \\ \hline
BBP (single para.) & 0.441       & 0.455       & 0.128       \\
BBP (full window)  & \textbf{0.284}       & \textbf{0.305}       & \textbf{0.453}       \\ \hline
\end{tabular}
\caption{Metrics for global chapter break insertion}
\label{tab:with_dp_eval}
\end{table}

\begin{figure*}
    \begin{minipage}[t]{.6\textwidth}
    \vspace{-36mm}
    \begin{tabular}{|r|rrr|rr|r|}
\hline
\textbf{Algorithm} & \textbf{MSE}    & \textbf{MAE}   & \textbf{R2}   & \textbf{Pk}   & \textbf{WD}   & \textbf{F1}   \\ \hline
Baseline (\# sent) & 205.97          & 8.928          & 0.44          & -             & -             & -             \\ \hline
WOC (win=50)       & 203.26          & 8.797          & 0.45          & 0.46          & 0.50          & 0.13          \\
WOC (win=100)      & 203.23          & 8.804          & 0.45          & 0.45          & 0.49          & 0.14          \\
WOC (win=150)      & 203.19          & 8.805          & 0.45          & 0.44          & 0.49          & 0.14          \\
WOC (win=200)      & 203.17          & 8.804          & 0.45          & 0.44          & 0.49          & 0.14          \\ \hline
BBP (thr=0.9)      & 192.22          & 8.366          & 0.48          & 0.33          & 0.38          & 0.41          \\
BBP (thr=0.99)     & \textbf{188.08} & \textbf{8.155} & \textbf{0.49} & \textbf{0.32} & \textbf{0.38} & \textbf{0.41} \\ \hline
\end{tabular}
    \captionof{table}
      {%
        Evaluation metrics for regression to predict number of chapter breaks in a book (Window size [WOC] and threshold [BBP] denoted in parentheses)%
        \label{tab:num_chap_regr}%
      }
  \end{minipage}
  \hspace{5mm}
  \begin{minipage}[t]{.35\textwidth}
    \centering
    \includegraphics[width=\linewidth]{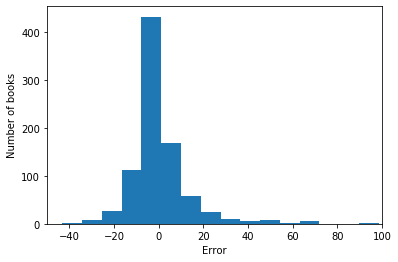}
    \caption
      {%
        Error distribution over test set using predictions for number of chapters from BBP (threshold=0.99)%
        \label{fig:error_dist_test_set}%
      }%
  \end{minipage}\hfill
  
\end{figure*}

Table \ref{tab:with_dp_eval} shows the evaluation metrics for global chapter break insertion. The dynamic programming approach consistently improves the WindowDiff and F1 metrics. The BERT model (full window) gives the best performance in terms of all three metrics.

\subsection{Estimating the Number of Breaks}

The models described above require the number of chapter boundaries to be specified. We now address the independent question of estimating how many chapter breaks to insert.

\begin{figure}[H]
  \begin{subfigure}[b]{0.495\linewidth}
    \includegraphics[width=\textwidth]{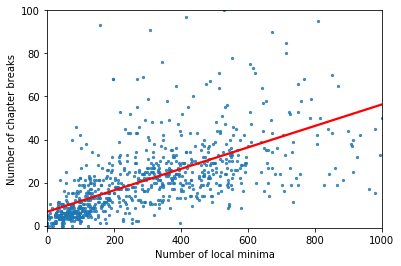}
    \caption{WOC local minima (window size $200$)\\ (Slope = $0.045$)}
    \label{fig:1}
  \end{subfigure}
  \begin{subfigure}[b]{0.495\linewidth}
    \includegraphics[width=\textwidth]{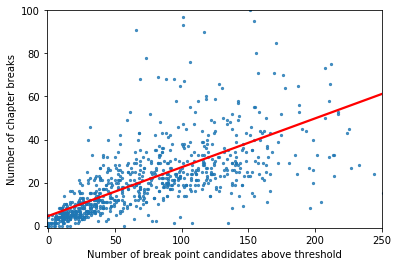}
    \caption{BBP (full-window) candidates (threshold = $0.99$)\\ (Slope = $0.213$)}
    \label{fig:2}
  \end{subfigure}
  \caption{Number of chapter breaks as a function of the number of candidate break points}
  \label{fig:num_chap_regr}
\end{figure}

Figure \ref{fig:num_chap_regr} shows the number of chapters against the number of break point candidates for both the approaches. The number of local minima in WOC are approximately 20 times the number of chapter breaks, reflecting potential event boundaries within chapters. The number of break point candidates obtained using BERT are approximately 5 times the number of chapter breaks. This can also be seen in Figures \ref{fig:woc_graph} and \ref{fig:bert_graph}. Although the BBP model performs better at exact break prediction, the WOC model provides more information in terms of events within chapters.

We now use a regression model to predict the number of breaks, with the number of candidate break points and the total number of sentences in the book, as features. For the number of candidate breaks, we use:
\begin{itemize}[topsep=5pt]
\itemsep0em 
    \item \textbf{WOC:} The total number of local minima 
    \item \textbf{BBP:} The number of candidate break points above a certain threshold.
\end{itemize}

We perform experiments on 2,626 books in the test set, so as to keep the results comparable for both the approaches.  We perform a train-test split of 67-33\%. We predict the number of chapter breaks using this regression, and further evaluate global break prediction with $\alpha=0.8$.

Table \ref{tab:num_chap_regr} shows the evaluation metrics on the test set, for regression using the models described above. The full-window BERT model shows the best performance in predicting the number of chapter breaks as well as break locations. Figure \ref{fig:error_dist_test_set} shows the error distribution over the test set for the best performing model.

\section{Conclusion and Future Work}

We build a chapter segmentation dataset resource consisting of 9,126 English fiction novels, using a hybrid approach combining neural inference and regular expression-based rule matching. We achieve and F1 score of $0.77$ on this task. Further, we use this dataset, remove structural cues, and address the task of predicting chapter boundaries. We present two methods for chapter segmentation. Our supervised approach achieves the best performance in exact break prediction, while our unsupervised approach provides information about potential sub-chapter break points.

Our work opens up avenues for further research in text segmentation, with potential applications in summarization and discourse analysis.
Potential future work includes combining the neural and cut-based approaches into a stronger method.
Finally, it would be interesting to do a deeper dive into variations of author strategies in chapterization, focusing more intently on books with large numbers of short chapters as being more reflective of episode boundaries.

\section*{Acknowledgments}
We thank the anonymous reviewers for their helpful feedback. This work was partially supported by NSF grants IIS-1926751, IIS-1927227, and IIS-1546113.


\bibliography{chaptercaptor}

\begin{thebibliography}{29}
\expandafter\ifx\csname natexlab\endcsname\relax\def\natexlab#1{#1}\fi

\bibitem[{Alemi and Ginsparg(2015)}]{alemi_2015}
Alexander~A Alemi and Paul Ginsparg. 2015.
\newblock {Text Segmentation based on Semantic Word Embeddings}.
\newblock \emph{arXiv preprint arXiv:1503.05543}.

\bibitem[{Badjatiya et~al.(2018)Badjatiya, Kurisinkel, Gupta, and
  Varma}]{badjatiya_2018}
Pinkesh Badjatiya, Litton~J Kurisinkel, Manish Gupta, and Vasudeva Varma. 2018.
\newblock Attention-based neural text segmentation.
\newblock In \emph{European Conference on Information Retrieval}, pages
  180--193. Springer.

\bibitem[{Barzilay and Malioutov(2006)}]{barzilay_2006}
Regina Barzilay and Igor Malioutov. 2006.
\newblock {Minimum Cut Model for Spoken Lecture Segmentation}.
\newblock In \emph{In Proceedings of the Annual Meeting of the Association for
  Computational Linguistics (COLING-ACL 2006}. Citeseer.

\bibitem[{Beeferman et~al.(1999)Beeferman, Berger, and
  Lafferty}]{beeferman_1999}
Doug Beeferman, Adam Berger, and John Lafferty. 1999.
\newblock {Statistical Models for Text Segmentation}.
\newblock \emph{Machine learning}, 34(1-3):177--210.

\bibitem[{Choi(2000)}]{choi_2000}
Freddy~YY Choi. 2000.
\newblock {Advances in Domain Independent Linear Text Segmentation}.
\newblock \emph{arXiv preprint cs/0003083}.

\bibitem[{Choi et~al.(2001)Choi, Wiemer-Hastings, and Moore}]{choi_2001}
Freddy~YY Choi, Peter Wiemer-Hastings, and Johanna~D Moore. 2001.
\newblock {Latent Semantic Analysis for Text Segmentation}.
\newblock In \emph{Proceedings of the 2001 conference on empirical methods in
  natural language processing}.

\bibitem[{D{\'e}jean and Meunier(2005)}]{dejean_2005}
Herv{\'e} D{\'e}jean and Jean-Luc Meunier. 2005.
\newblock {Structuring Documents According to Their Table of Contents}.
\newblock In \emph{Proceedings of the 2005 ACM symposium on Document
  engineering}, pages 2--9.

\bibitem[{D{\'e}jean and Meunier(2009)}]{dejean_2009}
Herv{\'e} D{\'e}jean and Jean-Luc Meunier. 2009.
\newblock On tables of contents and how to recognize them.
\newblock \emph{International Journal of Document Analysis and Recognition
  (IJDAR)}, 12(1):1--20.

\bibitem[{Devlin et~al.(2019)Devlin, Chang, Lee, and Toutanova}]{devlin_2019}
Jacob Devlin, Ming-Wei Chang, Kenton Lee, and Kristina Toutanova. 2019.
\newblock \href {https://doi.org/10.18653/v1/N19-1423} {{BERT}: Pre-training of
  deep bidirectional transformers for language understanding}.
\newblock In \emph{Proceedings of the 2019 Conference of the North {A}merican
  Chapter of the Association for Computational Linguistics: Human Language
  Technologies, Volume 1 (Long and Short Papers)}, pages 4171--4186,
  Minneapolis, Minnesota. Association for Computational Linguistics.

\bibitem[{Doucet et~al.(2013)Doucet, Kazai, Colutto, and
  M{\"u}hlberger}]{doucet_2013}
Antoine Doucet, Gabriella Kazai, Sebastian Colutto, and G{\"u}nter
  M{\"u}hlberger. 2013.
\newblock {Overview of the ICDAR 2013 Competition on Book Structure
  Extraction}.
\newblock In \emph{2013 12th International Conference on Document Analysis and
  Recognition}, pages 1438--1443. IEEE.

\bibitem[{Eisenstein(2009)}]{eisenstein_2009}
Jacob Eisenstein. 2009.
\newblock {Hierarchical Text Segmentation from Multi-scale Lexical Cohesion}.
\newblock In \emph{Proceedings of Human Language Technologies: The 2009 Annual
  Conference of the North American Chapter of the Association for Computational
  Linguistics}, pages 353--361.

\bibitem[{Eisenstein and Barzilay(2008)}]{eisenstein_2008}
Jacob Eisenstein and Regina Barzilay. 2008.
\newblock {Bayesian Unsupervised Topic Segmentation}.
\newblock In \emph{Proceedings of the 2008 Conference on Empirical Methods in
  Natural Language Processing}, pages 334--343.

\bibitem[{Glava{\v{s}} et~al.(2016)Glava{\v{s}}, Nanni, and
  Ponzetto}]{glavavs_2016}
Goran Glava{\v{s}}, Federico Nanni, and Simone~Paolo Ponzetto. 2016.
\newblock {Unsupervised Text Segmentation using Semantic Relatedness Graphs}.
\newblock Association for Computational Linguistics.

\bibitem[{Gutenberg(n.d.)}]{project_gutenberg}
Project Gutenberg. n.d.
\newblock \url{www.gutenberg.org}.
\newblock Accessed: May 2020.

\bibitem[{Hearst(1994)}]{hearst_1994}
Marti~A Hearst. 1994.
\newblock {Multi-Paragraph Segmentation of Expository Text}.
\newblock In \emph{Proceedings of the 32nd annual meeting on Association for
  Computational Linguistics}, pages 9--16. Association for Computational
  Linguistics.

\bibitem[{Joty et~al.(2019)Joty, Carenini, Ng, and Murray}]{discourse_2019}
Shafiq Joty, Giuseppe Carenini, Raymond Ng, and Gabriel Murray. 2019.
\newblock \href {https://doi.org/10.18653/v1/P19-4003} {Discourse analysis and
  its applications}.
\newblock In \emph{Proceedings of the 57th Annual Meeting of the Association
  for Computational Linguistics: Tutorial Abstracts}, pages 12--17, Florence,
  Italy. Association for Computational Linguistics.

\bibitem[{Kazantseva and Szpakowicz(2011)}]{kazantseva_2011}
Anna Kazantseva and Stan Szpakowicz. 2011.
\newblock {Linear Text Segmentation using Affinity Propagation}.
\newblock In \emph{Proceedings of the Conference on Empirical Methods in
  Natural Language Processing}, pages 284--293. Association for Computational
  Linguistics.

\bibitem[{Koshorek et~al.(2018)Koshorek, Cohen, Mor, Rotman, and
  Berant}]{koshorek_2018}
Omri Koshorek, Adir Cohen, Noam Mor, Michael Rotman, and Jonathan Berant. 2018.
\newblock {Text Segmentation as a Supervised Learning Task}.
\newblock \emph{arXiv preprint arXiv:1803.09337}.

\bibitem[{Manning et~al.(2014)Manning, Surdeanu, Bauer, Finkel, Bethard, and
  McClosky}]{stanford_corenlp}
Christopher~D. Manning, Mihai Surdeanu, John Bauer, Jenny Finkel, Steven~J.
  Bethard, and David McClosky. 2014.
\newblock \href {http://www.aclweb.org/anthology/P/P14/P14-5010} {The
  {Stanford} {CoreNLP} natural language processing toolkit}.
\newblock In \emph{Association for Computational Linguistics (ACL) System
  Demonstrations}, pages 55--60.

\bibitem[{McConnaughey et~al.(2017)McConnaughey, Dai, and
  Bamman}]{mcconnaughey_2017}
Lara McConnaughey, Jennifer Dai, and David Bamman. 2017.
\newblock {The Labeled Segmentation of Printed Books}.
\newblock In \emph{Proceedings of the 2017 Conference on Empirical Methods in
  Natural Language Processing}, pages 737--747.

\bibitem[{Mikolov et~al.(2013)Mikolov, Sutskever, Chen, Corrado, and
  Dean}]{mikolov_2013}
Tomas Mikolov, Ilya Sutskever, Kai Chen, Greg~S Corrado, and Jeff Dean. 2013.
\newblock Distributed representations of words and phrases and their
  compositionality.
\newblock In \emph{Advances in neural information processing systems}, pages
  3111--3119.

\bibitem[{Pevzner and Hearst(2002)}]{pevzner_2002}
Lev Pevzner and Marti~A Hearst. 2002.
\newblock {A Critique and Improvement of an Evaluation Metric for Text
  Segmentation}.
\newblock \emph{Computational Linguistics}, 28(1):19--36.

\bibitem[{Reynar(1994)}]{reynar_1994}
Jeffrey~C Reynar. 1994.
\newblock {An Automatic Method of Finding Topic Boundaries}.
\newblock In \emph{Proceedings of the 32nd annual meeting on Association for
  Computational Linguistics}, pages 331--333. Association for Computational
  Linguistics.

\bibitem[{Riedl and Biemann(2012)}]{riedl_2012}
Martin Riedl and Chris Biemann. 2012.
\newblock {Text Segmentation with Topic Models}.
\newblock \emph{Journal for Language Technology and Computational Linguistics},
  27(1):47--69.

\bibitem[{Sakahara et~al.(2014)Sakahara, Okada, and Nitta}]{sakahara_2014}
Makoto Sakahara, Shogo Okada, and Katsumi Nitta. 2014.
\newblock {Domain-independent Unsupervised Text Segmentation for Data
  Management}.
\newblock In \emph{2014 IEEE International Conference on Data Mining Workshop},
  pages 481--487. IEEE.

\bibitem[{Sims et~al.(2019)Sims, Park, and Bamman}]{sims_2019}
Matthew Sims, Jong~Ho Park, and David Bamman. 2019.
\newblock {Literary Event Detection}.
\newblock In \emph{Proceedings of the 57th Annual Meeting of the Association
  for Computational Linguistics}, pages 3623--3634.

\bibitem[{Utiyama and Isahara(2001)}]{utiyama_2001}
Masao Utiyama and Hitoshi Isahara. 2001.
\newblock {A Statistical Model for Domain-independent Text Segmentation}.
\newblock In \emph{Proceedings of the 39th Annual Meeting of the Association
  for Computational Linguistics}, pages 499--506.

\bibitem[{Wu et~al.(2013)Wu, Mitra, and Giles}]{wu_2013}
Zhaohui Wu, Prasenjit Mitra, and C~Lee Giles. 2013.
\newblock {Table of Contents Recognition and Extraction for Heterogeneous Book
  Documents}.
\newblock In \emph{2013 12th International Conference on Document Analysis and
  Recognition}, pages 1205--1209. IEEE.

\bibitem[{Yamron et~al.(1998)Yamron, Carp, Gillick, Lowe, and van
  Mulbregt}]{yamron_1998}
Jonathan~P Yamron, Ira Carp, Larry Gillick, Steve Lowe, and Paul van Mulbregt.
  1998.
\newblock {A Hidden Markov Model Approach to Text Segmentation and Event
  Tracking}.
\newblock In \emph{Proceedings of the 1998 IEEE International Conference on
  Acoustics, Speech and Signal Processing, ICASSP'98 (Cat. No. 98CH36181)},
  volume~1, pages 333--336. IEEE.

\end{thebibliography}
\bibliographystyle{acl_natbib}
\end{document}